\documentclass[journal]{IEEEtran}

\usepackage{amsmath,amssymb}
\usepackage{booktabs}
\usepackage{cite}
\usepackage{graphicx}
\usepackage{multirow}
\usepackage{xcolor}
\usepackage{url}
\usepackage[hidelinks]{hyperref}
\usepackage{placeins}
\usepackage{stfloats}
\usepackage{xspace}

\makeatletter
\renewcommand{\@IEEEfigurecaptionsepspace}{\vskip4pt\relax}
\makeatother

\graphicspath{{figures/}}
\newcommand{\method}{ReinSeg\xspace}

\title{ReinSeg: Retrieval-Guided Bounded Multimodal Refinement for 3-D Road Anomaly Segmentation}

\author{Haifa~Zhang, Yijing~Wang, and Zhiqiang~Zuo
\thanks{The authors are with the School of Electrical and Information Engineering, Tianjin University, Tianjin, China.}}

\hypersetup{pdftitle={ReinSeg: Retrieval-Guided Bounded Multimodal Refinement for 3-D Road Anomaly Segmentation},pdfauthor={Haifa Zhang, Yijing Wang, Zhiqiang Zuo},pdfsubject={Research preprint}}
\begin{document}
\maketitle
\thispagestyle{plain}
\pagestyle{plain}

\begin{abstract}
3-D road anomaly segmentation is essential for risk-aware perception in
autonomous driving. Its central challenge is to rank sparsely observed hazards
ahead of abundant normal road structures while preserving point-wise metric
localization. LiDAR provides accurate 3-D point support, whereas surround-view
imagery offers complementary semantic context for separating unexpected hazards
from challenging normal structures. We propose ReinSeg, a retrieval-guided
bounded multimodal refinement framework for point-wise road anomaly
segmentation. Occupied LiDAR bird's-eye-view cells retrieve scene-specific
context from a shared memory of eight surround-view images without supplying
calibration variables to the retrieval module. A relative-norm-bounded residual
returns this context to the LiDAR point features, so the frozen LIDO
anomaly-scoring interface is preserved, and the refinement is optimized with
in-distribution semantic labels only. On the Spotting the Unexpected (STU)
benchmark, ReinSeg improves validation average precision from 27.54\% to
35.67\% and reduces the false-positive rate at 95\% anomaly recall from
34.85\% to 10.61\%, shrinking the high-recall screening set from 67.54 to 20.55
million points. Relative to REL, AP is 24.99 percentage points higher and
FPR95 is 0.99 points lower. Matched interfaces and component isolation show that retrieved
scene imagery and the relative-norm bound are both essential, and ReinSeg stays
above LiDAR-only performance under missing, corrupted, or misaligned camera
inputs while running at 21.8 frames per second. These results establish bounded
retrieval of scene-specific visual evidence as an effective route to reliable
3-D road-hazard perception.
\end{abstract}


\section{Introduction}
\label{sec:introduction}

\IEEEPARstart{T}{hree}-dimensional (3-D) road anomaly segmentation enables autonomous vehicles to
identify previously unseen road hazards while preserving their metric
locations. Fallen cargo, detached vehicle parts, and unusual debris may occupy
only a small region and produce few LiDAR returns. Reliable point-wise
perception therefore requires ranking these sparse anomalous observations
ahead of millions of normal road points, including challenging road
boundaries, sidewalks, and terrain transitions. For safety-critical intelligent
transportation systems, risk-aware perception must recover such hazards at high
recall while keeping the number of normal points that a downstream monitor or
planner has to screen small.

The Spotting the Unexpected (STU) benchmark~\cite{nekrasov2025stu} establishes a
representative evaluation setting for 3-D road anomaly segmentation, providing
densely annotated LiDAR points, synchronized surround-view imagery, and
sequential observations of unexpected road objects. LiDAR-only anomaly rankers
preserve the metric point support and provide a direct interface for point-wise
hazard localization. LIDO~\cite{mosco2026lido}, for example, models known-class
feature distributions and scores individual LiDAR points without target-domain
anomaly supervision. On STU validation, LIDO still retains 67.54 million
in-distribution points at the threshold that recovers 95\% of anomaly points,
most of them on the road surface. This high-recall screening burden motivates
bringing complementary scene semantics into point-wise LiDAR anomaly ranking.

Surround-view cameras complement LiDAR with dense appearance and scene context.
Existing camera--LiDAR systems commonly associate the modalities through
calibrated projection, depth lifting, or a jointly learned prediction head
\cite{vora2020pointpainting,philion2020lift,liu2023bevfusion}. Point-wise
anomaly segmentation poses a different interface problem: LiDAR already defines
where predictions are evaluated, and the anomaly scorer already defines how point
features become anomaly scores in a carefully structured feature space.
Effective refinement must therefore answer two questions together: which visual
evidence is relevant to each occupied LiDAR region, and how far that evidence may
move each point feature while the scorer remains reliable.

We introduce \method, which couples visual evidence retrieval with point-wise
influence control. Occupied LiDAR bird's-eye-view (BEV) cells query a shared
token memory built from all eight surround-view images through
calibration-input-free retrieval. The selected context is returned to each point
through a residual whose norm is bounded by a fixed fraction of the LiDAR feature
norm, and the refined features are scored by the original LIDO function. Only the
visual retrieval and refinement modules are optimized, using in-distribution
semantic labels, while the LiDAR encoder and scorer remain frozen. \method thus
determines both what image evidence enters the LiDAR ranking pipeline and how
strongly it may change each point feature.

On STU, \method raises validation average precision from 27.54\% for LIDO to
35.67\% and reduces the false-positive rate at 95\% anomaly recall (FPR95) from
34.85\% to 10.61\%, shrinking the high-recall screening set from 67.54 to 20.55
million points. It also exceeds REL~\cite{li2026rel} by 24.99 AP points
while reducing FPR95 by 0.99 points. Against a frozen LiDAR reference with the same encoder and
scorer, the retrieved evidence adds 5.65 AP points and improves 16 of the 19
evaluation sequences. Controlled ablations attribute this gain to retrieved
scene imagery and to the relative-norm bound, and \method remains above the
LiDAR-only baseline under missing, corrupted, or misaligned camera inputs while
running at 21.8 frames per second.

The main contributions are as follows:

\begin{itemize}
    \item We formulate camera-assisted point-wise road anomaly segmentation
    through two coupled requirements: selecting scene-relevant visual evidence
    for each occupied LiDAR region and bounding its influence on the point
    features consumed by a frozen anomaly scorer.
    \item We develop \method, which combines calibration-input-free retrieval
    from a surround-view visual memory with relative-norm-bounded feature
    refinement. The framework preserves LiDAR-defined metric support and the
    original anomaly-scoring interface and is optimized with known-class labels
    only.
    \item We show on STU that \method substantially improves point-wise hazard
    ranking and reduces the high-recall false-alarm burden, and we verify through
    matched interfaces, component isolation, and sensor interventions that the
    gain stems from retrieved scene evidence, that the bound is essential, and
    that degraded cameras never push the system below LiDAR-only performance.
\end{itemize}

\section{Related Work}
\label{sec:related}

\subsection{Road Anomaly Segmentation}

Image-level out-of-distribution detection has developed confidence-, energy-,
and outlier-exposure-based scoring methods
\cite{hendrycks2017baseline,liang2018odin,liu2020energy,hendrycks2019oe,yang2024oodsurvey,wu2024tib}.
Dense road-scene benchmarks subsequently extended this objective to pixel-wise
localization of unexpected regions
\cite{pinggera2016lost,lis2019resynthesis,blum2021fishyscapes,chan2021segmentme,dibiase2021pixelwise,tian2022pebal,nayal2023rba}.
DetSeg adds object-level structure to this camera-only setting by detecting
open-world objects and temporally stabilizing the retained regions
\cite{zhu2025detseg}, and likelihood-ratio segmentation separates unknown regions
from closed-set predictions \cite{nayal2025likelihood}. Point-wise 3-D
perception additionally requires each anomaly score to retain the
metric location of an observed LiDAR return under extreme class imbalance.

SemanticKITTI established dense closed-set semantics for sequential LiDAR
perception \cite{behley2019semantickitti}, whereas STU provides synchronized
sensor data and dense point labels for real unexpected road objects
\cite{nekrasov2025stu}. LIDO models the feature geometry
of known semantic classes and combines semantic and contrastive cues to score
point-wise anomalies without real anomaly samples \cite{mosco2026lido}. REL
strengthens LiDAR anomaly learning through pseudo-outlier construction
\cite{li2026rel}. LIDO's known-class feature geometry and point-wise scoring
rule provide the reference interface for camera-assisted refinement in this
work.

\subsection{Multimodal 3-D Road-Hazard Perception}

Camera--LiDAR systems exploit complementary appearance and geometry for small
obstacle segmentation, unknown-object filtering, multimodal anomaly benchmarks,
and point-wise out-of-distribution detection
\cite{singh2020small,bogdoll2022unknown,bogdoll2024anovox,liu2025featuremixing}.
Detection architectures have explored point decoration, proposal or voxel
fusion, cross-view aggregation, and repeated modality interaction
\cite{li2022deepfusion,yang2022deepinteraction}.
Related occupancy methods further study fused or anomalous voxels in multimodal
road scenes \cite{ming2025occfusion,zhang2025occood,zhang2026proood}. Among
box-, voxel-, and point-level outputs, only the last ties every anomaly score
directly to a measured LiDAR return.

Recent T-ITS studies improve autonomous-driving perception through robust data
augmentation, compact multi-sensor learning, explicit LiDAR--camera alignment,
and task-specific cross-modal interaction
\cite{zheng2023robust,natan2022compact,zhu2024alignment,liu2025railfusion}.
Complementary work on point-cloud ground segmentation and camera-based semantic
occupancy further illustrates the importance of uncertainty-aware metric
reasoning and dense scene semantics \cite{li2026fugseg,fang2026dsocc}.

Multimodal industrial anomaly detection has shown that cross-modal retrieval
can expose evidence that is weak in a single modality
\cite{wang2023m3dm,costanzino2024crossmodal}. This retrieval perspective motivates the use of surround-view context to
enrich sparse LiDAR evidence in road scenes.

Table~\ref{tab:positioning} summarizes representative visual and 3-D anomaly
formulations through their output representations and association mechanisms.

\begin{table*}[t]
\caption{Positioning of representative visual and 3-D anomaly formulations.
``Assoc.'' denotes how visual evidence is associated with the 3-D
representation; ``bounded update'' denotes an explicit point-wise constraint on
the multimodal feature update.}
\label{tab:positioning}
\centering
\footnotesize
\setlength{\tabcolsep}{5.0pt}
\begin{tabular}{lllllc}
\toprule
Method & Scene & Output & Inference input & Assoc. & Bounded update \\
\midrule
M3DM \cite{wang2023m3dm} & Industrial & 3-D region & RGB + point cloud & Registered RGB--3-D & No \\
Unknown-road detection \cite{bogdoll2022unknown} & Road & 3-D object & RGB + LiDAR & Calibrated projection & No \\
DetSeg \cite{zhu2025detseg} & Road & 2-D object & RGB sequence & Temporal boxes & No \\
Feature Mixing \cite{liu2025featuremixing} & Road & 3-D point & RGB + LiDAR & Calibrated point fusion & No \\
OccOoD \cite{zhang2025occood} & Road & 3-D voxel & Multi-camera & Camera occupancy & No \\
LIDO \cite{mosco2026lido} & Road & 3-D point & LiDAR & n/a & No \\
\textbf{\method} & Road & 3-D point & RGB + LiDAR & Content retrieval & Yes \\
\bottomrule
\end{tabular}
\end{table*}

\subsection{Camera--LiDAR Fusion Interfaces}

Conventional fusion constructs an explicit spatial correspondence. Automotive
perception surveys frame this design space through what, when, and how sensor
features are combined \cite{feng2021multimodal}.
PointPainting projects image predictions onto LiDAR points \cite{vora2020pointpainting},
Lift-Splat-Shoot lifts perspective features through depth distributions
\cite{philion2020lift}, and BEVFusion combines independently encoded modalities
in a common BEV \cite{liang2022bevfusion,liu2023bevfusion}. Query-based fusion
such as TransFusion and CMT shows that attention can transfer complementary
evidence between sensor representations
\cite{bai2022transfusion,yan2023cmt}. Learned Fusion and CFT further show that
cross-modal or multi-camera mappings can be learned without supplying
calibration parameters to the fusion module \cite{furst2023learned,jiang2022cft}.

These systems suggest three fusion interfaces. Projection-based methods attach
image features to selected 3-D elements through known geometry. Lift-based
methods first construct a camera-derived spatial representation. Query-based
methods instead let one representation select information from another through
content-dependent weights. A query-based interface suits point-wise anomaly
ranking because occupied LiDAR cells already define the metric query locations.
\method adopts it and adds what anomaly scoring requires: an explicit bound on
how far the retrieved evidence may move each point feature before the original
scorer is applied.

\section{Method}
\label{sec:method}

\subsection{Problem Formulation and Design Principles}

\method is a retrieval-guided bounded multimodal refinement framework built
around the point-wise LIDO anomaly scorer. The evaluated LiDAR points remain the
query and output domain, while images refine the intermediate point features
before the original LiDAR scoring rule is applied.

For one synchronized sample, the input is a LiDAR point cloud
$\mathcal{P}=\{(\mathbf{x}_i,\mathbf{r}_i)\}_{i=1}^{N_p}$ and eight RGB images
$\{\mathbf{I}_v\}_{v=1}^{8}$, where $\mathbf{x}_i\in\mathbb{R}^{3}$ is a metric
coordinate and $\mathbf{r}_i$ contains the remaining LiDAR attributes. The model
assigns a continuous anomaly score $s_i\in[0,1]$ to every evaluated point.
During optimization, $\mathcal{K}\subseteq\{1,\ldots,N_p\}$ denotes the points
with in-distribution semantic labels, hereafter called ID points. Target-domain
anomaly annotations are used only for final evaluation.

\begin{figure*}[t]
    \centering
    \includegraphics[width=\textwidth]{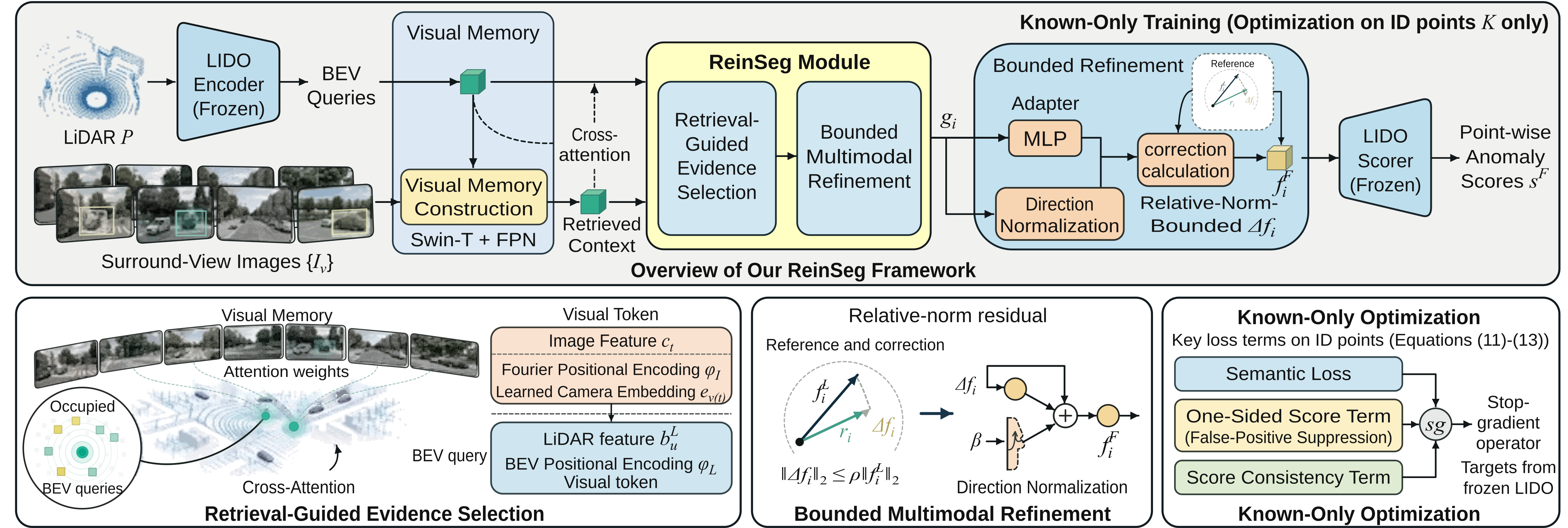}
    \caption{Overview of \method. Occupied LiDAR BEV queries retrieve
    scene-relevant context from an eight-view visual memory. The retrieved
    evidence is returned through a relative-norm-bounded residual, known-only
    objectives optimize the visual path, and the frozen LIDO scorer preserves
    the point-wise anomaly-scoring interface.}
    \label{fig:architecture}
\end{figure*}

The framework couples two stages. Retrieval-guided evidence selection obtains
scene-specific visual context for occupied LiDAR BEV cells. Bounded multimodal
refinement returns this context to each point through a residual whose norm is
limited relative to the LiDAR feature norm. Together, they determine what visual
evidence enters the point-wise ranking pipeline and how strongly it may change
the LiDAR representation. Figure~\ref{fig:architecture} maps both stages onto one
forward path, and the complete scan-level mapping is

\begin{equation}
    \begin{aligned}
    \mathbf{s}^{F}&=g_{\mathrm{LIDO}}\!\left(
    \mathbf{F}^{L}_{p}+\Delta\mathbf{F}
    (\mathcal{P},\{\mathbf{I}_v\}_{v=1}^{8};\theta)\right),\\
    \|\Delta\mathbf{f}_i\|_2&\leq \rho\|\mathbf{f}_i^L\|_2,
    \end{aligned}
    \label{eq:method-overview}
\end{equation}

where $\mathbf{F}^{L}_{p}$ stacks the reference point features and
$\Delta\mathbf{F}$ is the learned multimodal correction, whose point-wise norm
never exceeds the fraction $\rho$ of the LiDAR feature norm.

\subsection{LiDAR Reference Scoring Interface}
\label{sec:reference}

The LiDAR reference interface follows LIDO's point representation and scoring
rule. The encoder $E_L$ uses sparse convolutions \cite{choy2019minkowski} to map
the voxelized point cloud to
$\mathbf{F}^{L}_{v}=E_L(\mathcal{P})\in\mathbb{R}^{N_v\times96}$, where $N_v$ is
the number of non-empty voxels. The voxel index $\nu(i)$ assigns each evaluated
point the feature $\mathbf{f}^{L}_i=\mathbf{F}^{L}_{v,\nu(i)}\in\mathbb{R}^{96}$
of its voxel, and every output score is written back to the original point
index, so the evaluated point set of the STU evaluator is unchanged. The LIDO
semantic and contrastive projections, class mean activation vectors, and anomaly
equation are denoted together by $g_{\mathrm{LIDO}}$, giving the reference scores

\begin{equation}
    \mathbf{s}^{L}=g_{\mathrm{LIDO}}(\mathbf{F}^{L}_{p}),
    \qquad s_i^L=[\mathbf{s}^L]_i .
    \label{eq:anchor-score}
\end{equation}

\textit{Reference adaptation.} Before refinement, $E_L$ and the LIDO projections
are adapted to the STU training scans, starting from the public LIDO weights.
A first stage trains them jointly with the visual path of
Sections~\ref{sec:retrieval}--\ref{sec:refinement} under LIDO's known-class
objectives, i.e., weighted cross-entropy, Lov\'asz, class-prototype,
contrastive, and objectosphere losses, distilled from the public model. A second
stage sharpens the reference ranking with class-conditional virtual outliers
sampled from the low-likelihood tails of the ID feature distributions
\cite{du2022vos} through an auxiliary outlier head that is discarded afterwards.
Both stages use ID labels only. The adapted $E_L$ and $g_{\mathrm{LIDO}}$ are
then frozen as the reference interface, and the visual path is initialized from
the first stage.

To form metric queries, voxel features are averaged along height and scattered
onto a $400\times400$ BEV grid with 0.25-m cells. For an occupied cell $u$ with
voxel set $\mathcal{V}(u)$, the LiDAR BEV feature is

\begin{equation}
    \mathbf{b}^{L}_u=\frac{1}{|\mathcal{V}(u)|}
    \sum_{j\in\mathcal{V}(u)}\mathbf{F}^{L}_{v,j}\in\mathbb{R}^{96},
    \qquad u\in\mathcal{U},
    \label{eq:bev-aggregation}
\end{equation}

where $\mathcal{U}$ contains the $M$ occupied cells and $\mathbf{p}_u\in[-1,1]^2$
is the normalized cell coordinate. Empty cells are omitted, so the query count
follows the occupied LiDAR support rather than the area of the BEV canvas.

\subsection{Calibration-Input-Free Visual Retrieval}
\label{sec:retrieval}

The retrieval module maps all eight images to a shared visual memory and lets
occupied BEV cells select scene-relevant content. It receives no camera
intrinsics, camera--LiDAR extrinsics, ego-pose transforms, rays, depth lifts, or
projected point coordinates; camera identity and image position organize the
visual tokens of the fixed synchronized rig.

One shared Swin-T encoder \cite{liu2021swin}, initialized from ImageNet-1K,
provides stages at strides 8, 16, and 32. A feature pyramid network (FPN)
\cite{lin2017fpn} maps them to 64 channels with $1\times1$ lateral convolutions,
upsamples them to stride 8, sums them, and applies a $3\times3$ output
convolution. The fused map is adaptively pooled to an $8\times14$ grid, which
preserves the image aspect ratio and yields 112 tokens per view and $T=896$
tokens in total. For token $t$ of camera $v(t)$ at normalized image position
$\boldsymbol{\xi}_t$,

\begin{equation}
    \mathbf{m}_t=\mathbf{W}_{I}\mathbf{c}_t
    +\phi_I(\boldsymbol{\xi}_t)+\mathbf{e}_{v(t)},
    \label{eq:image-token}
\end{equation}

where $\mathbf{c}_t$ is the pooled image feature, $\phi_I$ is a two-dimensional
Fourier position encoding, and $\mathbf{e}_{v(t)}$ is a learned camera-identity
embedding. Stacking all tokens gives $\mathbf{M}_I\in\mathbb{R}^{T\times64}$.
Each occupied cell forms the query
$\mathbf{q}_u=\mathbf{W}_{L}\mathbf{b}^{L}_u+\phi_L(\mathbf{p}_u)$ with a BEV
Fourier position encoding $\phi_L$. Four-head scaled dot-product cross-attention
\cite{vaswani2017attention} retrieves visual context from the complete memory,

\begin{equation}
    \mathbf{o}^{(k)}_u=
    \operatorname{softmax}\!\left(
    \frac{(\mathbf{q}_u\mathbf{W}^{(k)}_Q)
    (\mathbf{M}_I\mathbf{W}^{(k)}_K)^{\!\top}}{\sqrt{d_k}}
    \right)\mathbf{M}_I\mathbf{W}^{(k)}_V,
    \label{eq:attention-head}
\end{equation}

with $k\in\{1,\ldots,4\}$ and $d_k=16$, where the softmax runs over the $T$ image
tokens. The concatenated heads are projected, normalized, and refined by a
feed-forward network,

\begin{equation}
    \begin{aligned}
        \widehat{\mathbf{h}}_u
        &=\operatorname{LN}\!\left(
        [\mathbf{o}^{(1)}_u;\ldots;\mathbf{o}^{(4)}_u]\mathbf{W}_O\right),\\
        \widetilde{\mathbf{b}}^{I}_u
        &=\mathbf{W}_{R}\operatorname{LN}\!\left(\widehat{\mathbf{h}}_u+
        \operatorname{FFN}(\widehat{\mathbf{h}}_u)\right)\in\mathbb{R}^{64}.
    \end{aligned}
    \label{eq:retrieval}
\end{equation}

Each occupied cell thus compares its LiDAR content against all camera slots and
image regions and aggregates the evidence relevant to it into a cell-specific
visual representation.

The retrieved features and the LiDAR BEV features are scattered to
$\widetilde{\mathbf{B}}^I$ and $\mathbf{B}^L$ on the same grid, with zeros at
unoccupied cells, and a convolutional fuser exploits local BEV neighborhoods,

\begin{equation}
    \mathbf{G}=G_\psi\!\left([\widetilde{\mathbf{B}}^{I};\mathbf{B}^{L}]\right),
    \qquad \mathbf{g}_u=\mathbf{G}[u]\in\mathbb{R}^{96}.
    \label{eq:bev-fuser}
\end{equation}

$G_\psi$ applies a $3\times3$ convolution with batch normalization and ReLU to
the 160-channel concatenation, followed by a residual block of two $3\times3$,
96-channel convolutions. Each point gathers the fused feature of the cell
$\mu(i)$ that contains its voxel, $\mathbf{g}_i=\mathbf{g}_{\mu(i)}$, so points
in one cell share visual context while keeping their distinct LiDAR features.

\subsection{Bounded Multimodal Refinement}
\label{sec:refinement}

Bounded multimodal refinement regulates how strongly retrieved evidence changes
each LiDAR feature. A two-layer adapter $h_\theta$ with a 96-D hidden layer,
layer normalization, and GELU activation proposes a correction direction from
the reference feature and its retrieved context,
$\boldsymbol{\delta}_i=h_\theta([\mathbf{f}^{L}_i;\mathbf{g}_i])\in\mathbb{R}^{96}$.
Its magnitude is set relative to the reference feature through the budget
utilization $\beta=\rho\tanh(a)$:

\begin{equation}
    \Delta\mathbf{f}_i=
    \beta\,\|\mathbf{f}^{L}_i\|_2
    \frac{\boldsymbol{\delta}_i}
    {\max(\|\boldsymbol{\delta}_i\|_2,10^{-3})},
    \qquad
    \mathbf{f}^{F}_i=\mathbf{f}^{L}_i+\Delta\mathbf{f}_i .
    \label{eq:bounded-residual}
\end{equation}

Because $|\tanh(a)|<1$, every correction satisfies

\begin{equation}
    \|\Delta\mathbf{f}_i\|_2 \leq \rho\,\|\mathbf{f}^{L}_i\|_2
    \label{eq:bound}
\end{equation}

for any point, scene, and camera condition. We use $\rho=0.10$. The scalar $a$
is initialized to zero, so the first forward pass reproduces the reference
features exactly while $\partial\beta/\partial a=\rho$ activates the residual
path from the first update; $\beta/\rho$ reports the budget utilization. The
refined features $\mathbf{F}^{F}_{p}$ are scored by the unchanged LIDO function,

\begin{equation}
    \mathbf{s}^{F}=g_{\mathrm{LIDO}}(\mathbf{F}^{F}_{p}),
    \qquad s_i^F=[\mathbf{s}^F]_i .
    \label{eq:fused-score}
\end{equation}

The bound keeps the refined features in the neighborhood of the feature geometry
in which the class mean activation vectors and the anomaly equation are defined.
Visual evidence can therefore re-rank difficult normal structures and sparse
hazards without overriding the metric LiDAR representation on which the scorer
relies.

\subsection{Known-Only Optimization}

Anomaly-label-free optimization updates only the visual path: the image encoder,
retrieval block, BEV fuser, residual adapter, and $a$. With $\mathbf{z}^{F}_i$ the
semantic logits of the refined features under the LIDO semantic projection, the
semantic objective is

\begin{equation}
    \mathcal{L}_{\mathrm{sem}}=
    \operatorname{CE}_{w}(\mathbf{z}^{F}_{\mathcal K},\mathbf{y}_{\mathcal K})
    +1.5\,\mathcal{L}_{\mathrm{Lov}}
    (\mathbf{z}^{F}_{\mathcal K},\mathbf{y}_{\mathcal K}),
    \label{eq:semantic-loss}
\end{equation}

where $\operatorname{CE}_w$ uses LIDO's inverse-frequency class weights and
$\mathcal{L}_{\mathrm{Lov}}$ is the Lov\'asz--Softmax loss
\cite{berman2018lovasz}. A one-sided term discourages increased anomaly
responses on ID points, and a consistency term keeps refined ID scores close to
the reference ranking:

\begin{equation}
    \begin{aligned}
    \mathcal{L}_{\mathrm{fp}}&=
    \frac{1}{|\mathcal{K}|}\sum_{i\in\mathcal{K}}
    \max\!\left(0,s_i^{F}-\operatorname{sg}(s_i^{L})\right),\\
    \mathcal{L}_{\mathrm{cons}}&=
    \frac{1}{|\mathcal{K}|}\sum_{i\in\mathcal{K}}
    \operatorname{SmoothL1}\!\left(s_i^{F},\operatorname{sg}(s_i^{L})\right),
    \end{aligned}
    \label{eq:score-losses}
\end{equation}

where $\operatorname{sg}$ denotes stop-gradient. The complete objective is

\begin{equation}
    \mathcal{L}=\mathcal{L}_{\mathrm{sem}}
    +2\mathcal{L}_{\mathrm{fp}}+0.1\mathcal{L}_{\mathrm{cons}}.
    \label{eq:total-loss}
\end{equation}

Gradients flow through $\mathbf{f}^{F}_i$ to the adapter, fuser, retrieval block,
and image encoder, while the LiDAR encoder and the scoring interface stay frozen.
The semantic term organizes ID points with the retrieved appearance, and the two
score terms hold ID responses at or below the reference level. The unchanged
scorer then operates on a clearer ID representation, which reduces normal-point
competition near the top of the anomaly ranking.

\section{Experiments}
\label{sec:experiments}
\suppressfloats[t]

\subsection{Protocol and Metrics}

We evaluate on STU~\cite{nekrasov2025stu}, using synchronized high-resolution
LiDAR and all eight surround-view cameras. The two released known-only
sequences, 201 and 206, are divided by a fixed frame-index rule: the 227 scans
whose index is a multiple of five form the model-selection split, and the
remaining 904 scans form the training split. The
official STU anomaly validation split is the held-out evaluation set. We follow
the released validity and ignore masks, retain points from 2.5 to 50~m, and
evaluate 1,960 scans containing 193,879,969 points, of which 87,499 are
anomalous.

Average precision (AP) is the primary metric because it emphasizes the top of
the highly imbalanced ranking. AUROC measures global anomaly--ID ordering, and
FPR95 is the fraction of ID points retained at 95\% anomaly recall, which we
also report as the absolute ID-point screening burden. Scores are pooled over all
valid points and evaluated with the official STU point-level evaluator. Known-class
mIoU is measured on the 227 selection scans. To isolate the visual path, we also
report the frozen LiDAR reference of Eq.~\eqref{eq:anchor-score}, which shares
the encoder and scorer of \method and differs only by the absence of retrieved
visual evidence. Per-sequence AP is computed independently on each of the 19
evaluation sequences, and its mean change is bootstrapped over sequences so that
a long sequence cannot dominate the interval.

\subsection{Implementation Details}

LiDAR points are voxelized at 5~cm, and the $100\times100$~m BEV uses 0.25-m
cells. Images are resized to $256\times448$ without geometric augmentation. The
visual path is trained for 30 epochs with AdamW, batch size 4, automatic mixed
precision, weight decay $10^{-4}$, and global-gradient clipping at 5.0. The
retrieval, fusion, and adapter modules use a learning rate of $10^{-4}$; the Swin
stages are frozen for the first five epochs and then updated at $5\times10^{-7}$.
The checkpoint with the lowest known-only selection loss,
Eq.~\eqref{eq:total-loss} on the selection scans, is evaluated once on the
held-out split. The reference adaptation stages each run for 128 epochs. All
experiments use a single NVIDIA GeForce RTX 4090 GPU, and every ablation shares
the training budget, initialization, and checkpoint rule of the full model.

\subsection{Hazard Prioritization}

Table~\ref{tab:main-comparison} compares \method with LiDAR-only anomaly
segmentation methods on STU validation. \method achieves the strongest point
ranking, raising AP from 27.54\% for LIDO to 35.67\% and reducing FPR95 from
34.85\% to 10.61\%. Compared with REL, which trains with synthesized auxiliary
outliers, AP increases from 10.68\% to 35.67\% (+24.99 percentage points),
AUROC increases from 97.12\% to 97.24\% (+0.12 points), and FPR95 decreases
from 11.60\% to 10.61\% ($-0.99$ points).

\begin{table*}[t]
\caption{Point-wise anomaly segmentation on STU validation (\%). Results of the
M4F3D and MinkNet variants are reported in~\cite{mosco2026lido} and those of REL
in~\cite{li2026rel}; LIDO and \method are evaluated on the identical point
support with the official STU evaluator. Best results are bold.
$^{\dagger}$Feature-space virtual
outliers are used only in the reference adaptation of the LiDAR interface
(Section~\ref{sec:reference}); the multimodal refinement uses ID labels only.}
\label{tab:main-comparison}
\centering
\footnotesize
\setlength{\tabcolsep}{6.0pt}
\begin{tabular*}{\textwidth}{@{\extracolsep{\fill}}llccc@{}}
\toprule
Method & Input & AP $\uparrow$ & AUROC $\uparrow$ & FPR95 $\downarrow$ \\
\midrule
\multicolumn{5}{l}{\textit{LiDAR only; no auxiliary-OOD supervision}} \\
M4F3D + MC Dropout~\cite{yilmaz2024mask4former} & LiDAR & 0.17 & 65.76 & 79.82 \\
M4F3D + RbA~\cite{yilmaz2024mask4former} & LiDAR & 1.64 & 73.00 & 100.00 \\
M4F3D + Max Logit~\cite{yilmaz2024mask4former} & LiDAR & 2.02 & 87.27 & 68.76 \\
M4F3D + Void Classifier~\cite{yilmaz2024mask4former} & LiDAR & 2.62 & 89.77 & 79.50 \\
M4F3D + Deep Ensemble~\cite{yilmaz2024mask4former} & LiDAR & 6.94 & 90.93 & 37.34 \\
MinkNet + RbA~\cite{choy2019minkowski} & LiDAR & 0.05 & 53.93 & 92.07 \\
MinkNet + Void Classifier~\cite{choy2019minkowski} & LiDAR & 0.11 & 70.24 & 100.00 \\
MinkNet + Max Logit~\cite{choy2019minkowski} & LiDAR & 0.97 & 90.92 & 43.49 \\
MinkNet + MC Dropout~\cite{choy2019minkowski} & LiDAR & 1.41 & 91.98 & 35.64 \\
MinkNet + Deep Ensemble~\cite{choy2019minkowski} & LiDAR & 2.72 & 93.32 & 29.67 \\
LIDO~\cite{mosco2026lido} & LiDAR & 27.54 & 95.05 & 34.85 \\
\midrule
\multicolumn{5}{l}{\textit{LiDAR only; synthesized auxiliary-OOD training}} \\
REL~\cite{li2026rel} & LiDAR & 10.68 & 97.12 & 11.60 \\
\midrule
\multicolumn{5}{l}{\textit{LiDAR with surround-view imagery}} \\
\textbf{\method (ours)}$^{\dagger}$ & LiDAR + 8 RGB & \textbf{35.67} & \textbf{97.24} & \textbf{10.61} \\
\bottomrule
\end{tabular*}
\end{table*}

\begin{figure*}[t]
    \centering
    \includegraphics[width=\textwidth]{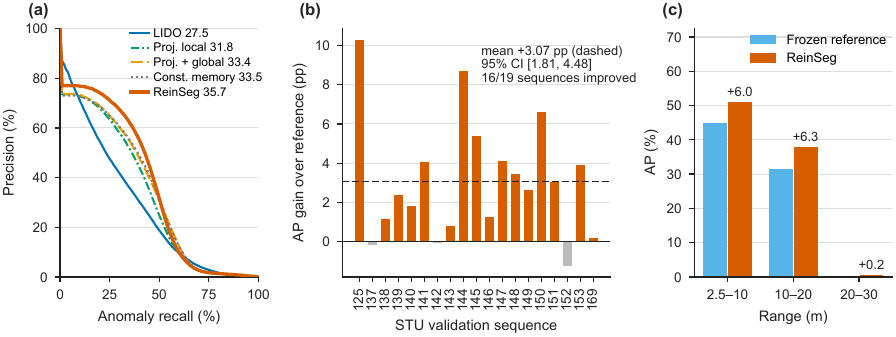}
    \caption{Operational performance on STU validation. (a) Point
    precision--recall of LIDO, the calibrated projection interfaces, constant
    visual memory, and \method; legend values are AP (\%). (b) Per-sequence AP
    gain of \method over the frozen LiDAR reference, which shares its encoder and
    scorer; the dashed line marks the mean. (c) AP by object range for the frozen
    reference and \method.}
    \label{fig:operational-performance}
\end{figure*}

The gain is produced by the retrieved visual evidence. Against the frozen
reference, which uses the same encoder and scorer without images, \method raises
AP from 30.02\% to 35.67\%, AUROC from 97.09\% to 97.24\%, and known-class mIoU
from 47.62\% to 53.42\%, while FPR95 falls from 12.87\% to 10.61\%. The
improvement is consistent across the evaluation data
(Fig.~\ref{fig:operational-performance}): the mean per-sequence AP gain is 3.07
points with a sequence-bootstrap 95\% interval of $[1.81,4.48]$, 16 of the 19
sequences improve, and every range bin gains, with $+6.0$ and $+6.3$ AP points
at 2.5--10~m and 10--20~m.

Lower FPR95 directly reduces the number of normal points that must be screened
before most hazards are found. At 95\% anomaly recall, LIDO retains 67.54
million ID points, the frozen reference 24.94 million, and \method 20.55 million.
Road-surface points dominate this burden and fall from 65.52 to 19.84 million,
so the high-recall operating regime of a safety monitor becomes far more
selective.

The matched interfaces in Table~\ref{tab:fusion-interface} isolate how image
evidence should reach the LiDAR scorer. Projected local bilinearly samples FPN
features at the projections of each point into the two calibrated front cameras,
after positive-depth, in-frame, and nearest-depth tests, and pools them into the
occupied BEV cells. Projected + global appends a pooled eight-view context token
to that local evidence. All interfaces share the visual backbone, BEV fuser,
relative-norm bound, losses, initialization, and training budget, and differ only
in how cells obtain visual context. Calibration-free retrieval yields the
highest AP, exceeding projected local and projected + global evidence by 3.91 and
2.29 points, respectively.

\begin{table}[t]
\caption{Matched visual interfaces on STU validation (\%). All variants share
the bounded refinement and training budget of \method.}
\label{tab:fusion-interface}
\centering
\footnotesize
\setlength{\tabcolsep}{3.6pt}
\begin{tabular}{lcccc}
\toprule
Interface & Calib. & AP $\uparrow$ & AUROC $\uparrow$ & FPR95 $\downarrow$ \\
\midrule
LIDO (no images) & n/a & 27.54 & 95.05 & 34.85 \\
Projected local & Yes & 31.76 & 97.35 & 9.96 \\
Projected + global & Yes & 33.38 & 97.26 & 10.18 \\
\textbf{\method retrieval} & \textbf{No} & \textbf{35.67} & 97.24 & 10.61 \\
\bottomrule
\end{tabular}
\end{table}

\subsection{Mechanism Analysis of Retrieval and Bounded Refinement}

Table~\ref{tab:component-ablation} traces where the improvement originates. All
rows share the frozen reference, training budget, initialization, and checkpoint
rule. The LiDAR adapter control trains the same fuser and adapter without image
context; constant visual memory replaces image tokens with learned
scene-independent tokens; global pooled RGB broadcasts one pooled eight-view
vector to every occupied cell without retrieval; and the unbounded row applies
$\Delta\mathbf{f}_i=\tanh(a)\,\boldsymbol{\delta}_i$ in place of
Eq.~\eqref{eq:bounded-residual}.

\begin{table}[t]
\caption{Component isolation on STU validation (\%). Best AP and FPR95 values are bold.}
\label{tab:component-ablation}
\centering
\footnotesize
\setlength{\tabcolsep}{2.4pt}
\begin{tabular}{lccccc}
\toprule
Variant & RGB & Ret. & Bound & AP $\uparrow$ & FPR95 $\downarrow$ \\
\midrule
Frozen LiDAR reference & -- & -- & -- & 30.02 & 12.87 \\
LiDAR adapter control & -- & -- & \checkmark & 32.99 & 11.58 \\
Constant visual memory & -- & \checkmark & \checkmark & 33.48 & 10.71 \\
Global pooled RGB & \checkmark & -- & \checkmark & 32.76 & 11.43 \\
Retrieval, unbounded & \checkmark & \checkmark & -- & 18.43 & 77.77 \\
Retrieval + bound + score gate & \checkmark & \checkmark & \checkmark & 32.02 & 12.33 \\
\textbf{\method} & \checkmark & \checkmark & \checkmark & \textbf{35.67} & \textbf{10.61} \\
\bottomrule
\end{tabular}
\end{table}

Retrieved scene imagery is the decisive ingredient. With the same bounded
interface, retrieving real surround-view content raises AP by 2.68 points over the
image-free adapter, by 2.19 points over learned scene-independent memory, and by
2.91 points over a globally pooled image vector. FPR95 also drops by 0.97, 0.10, and 0.82 points, respectively. Content that matches the current
scene and is selected per occupied cell therefore carries the information that
re-ranks hazards. The relative-norm bound is equally essential: removing it lets
the residual override the LiDAR representation, so AP collapses to 18.43\% and
FPR95 rises to 77.77\%. Scaling the bounded residual additionally by the reference
anomaly score restricts the correction to points that the reference already
ranks high and lowers AP to 32.02\%; applying the bounded correction to every
point lets retrieved context also reorganize the normal structures that compete
with hazards.

Table~\ref{tab:objective-structure} examines the training objectives and the
frozen scoring interface. The complete objective achieves the highest AP, and
removing the convolutional fuser lowers AP by 2.54 points, confirming the value
of local BEV context. Unfreezing the LIDO encoder and scorer converts the visual
path into a semantic segmentation refinement: mIoU rises to 79.34\%, but the
anomaly equation loses its calibrated ID feature geometry, AUROC drops to
91.00\%, and FPR95 increases to 51.81\%. Freezing the reference interface is
therefore what keeps the high-recall operating point selective.

\begin{table}[t]
\caption{Objective, fusion, and freezing controls on STU validation (\%).
``All'' denotes Eq.~\eqref{eq:total-loss}.}
\label{tab:objective-structure}
\centering
\footnotesize
\setlength{\tabcolsep}{3.4pt}
\begin{tabular}{lccc}
\toprule
Variant & AP $\uparrow$ & FPR95 $\downarrow$ & mIoU $\uparrow$ \\
\midrule
$\mathcal{L}_{\rm sem}$ only & 34.62 & 10.93 & 53.45 \\
$\mathcal{L}_{\rm sem}+\mathcal{L}_{\rm fp}$ & 34.90 & 10.82 & 53.40 \\
$\mathcal{L}_{\rm sem}+\mathcal{L}_{\rm cons}$ & 33.88 & 10.68 & 53.45 \\
All; no BEV fuser & 33.13 & 10.60 & 53.35 \\
All; LIDO unfrozen & 39.98 & 51.81 & 79.34 \\
\textbf{All; LIDO frozen} & \textbf{35.67} & \textbf{10.61} & 53.42 \\
\bottomrule
\end{tabular}
\end{table}

\begin{figure*}[t]
    \centering
    \includegraphics[width=\textwidth]{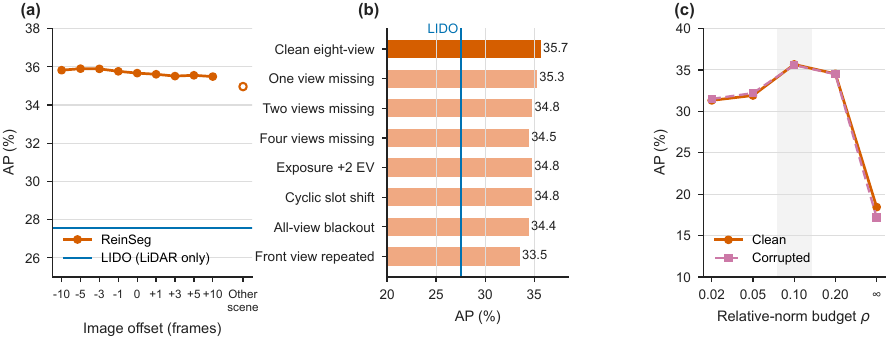}
    \caption{Mechanism and robustness analysis on STU validation. (a) AP under
    image--LiDAR offsets and with images from another scene of the same drive
    (open marker); the blue line is LiDAR-only LIDO. (b) Camera interventions at
    fixed weights. (c) Relative-norm budget $\rho$ on clean and corrupted
    images; the shaded column marks the adopted $\rho=0.10$.}
    \label{fig:causality-robustness}
\end{figure*}

We further probe the trained model with sensor-input interventions at fixed
weights (Fig.~\ref{fig:causality-robustness}). Temporal offsets shift the full
synchronized image set by up to $\pm10$ frames relative to LiDAR, and
sequence-local shuffling supplies all eight views from another frame of the same
drive at least 5~s and 25~m away. Camera tests apply balanced missing-view
masks, $+2$-EV exposure, a cyclic shift of camera slots, all-view blackout, and a
repeated front view. The budget sweep retrains the model for each $\rho$ and
evaluates it on clean images and under composite corruption, the mean over
four-view dropout, Gaussian blur $\sigma=5$, and $+2$-EV exposure.

The interventions connect the gain to scene-specific visual evidence. Offsets of
up to one second keep the images within the same scene and leave AP between
35.49\% and 35.90\%, whereas images from another scene of the same drive lower AP
to 34.96\%. Removing one, two, or four views lowers AP monotonically to 35.28\%,
34.80\%, and 34.46\%, and repeating the front view in all slots yields 33.52\%,
so the full surround-view memory contributes beyond any single camera. Under
every camera intervention, including complete blackout, \method stays 6.0--7.7 AP
points above LiDAR-only LIDO; a degraded camera input thus never reduces the
system below its LiDAR-only counterpart.

The relative-norm budget governs this behavior (Fig.~\ref{fig:causality-robustness}(c)).
AP rises from 31.33\% at $\rho=0.02$ and 31.92\% at $\rho=0.05$ to 35.67\% at
$\rho=0.10$, decreases to 34.54\% at $\rho=0.20$, and collapses to 18.43\% without a
bound. Under composite corruption the adopted budget retains 35.58\% AP, against
17.18\% for unbounded refinement. The learned utilization is $\beta/\rho=0.56$,
i.e., every correction uses 5.6\% of the reference feature norm. A small,
explicitly bounded update is thus sufficient to inject the selected visual
evidence while keeping the native LiDAR representation intact.

\subsection{Screening Behavior and Efficiency}

Table~\ref{tab:efficiency} positions \method on the accuracy--cost frontier. The
full eight-camera model runs at 45.9~ms per scan (21.8 frames/s) with 0.93~GiB of
inference memory, 15.0~ms above LiDAR-only LIDO. Reducing the memory to 56 tokens
per view retains 35.34\% AP. A stride-8 ResNet-18 image encoder lowers the
trainable parameters to 1.07~M and the latency to 41.0~ms while retaining 34.20\%
AP, and four cameras reach 32.12\%. Every configuration remains well above
LiDAR-only ranking, and the 112-token, eight-camera configuration is the
operating point used throughout the paper.

\begin{table*}[t]
\caption{Accuracy--efficiency trade-off at batch size one on one NVIDIA GeForce
RTX 4090 GPU. Best AP and FPR95 values are bold.}
\label{tab:efficiency}
\centering
\footnotesize
\setlength{\tabcolsep}{3.6pt}
\begin{tabular}{lcccccccc}
\toprule
Configuration & Cameras & Tokens/view & Trainable (M) & AP $\uparrow$ & FPR95 $\downarrow$
& Latency (ms) $\downarrow$ & FPS $\uparrow$ & Memory (GiB) $\downarrow$ \\
\midrule
LIDO & 0 & -- & -- & 27.54 & 34.85 & 30.9 & 32.3 & 0.52 \\
\method, 4 cameras & 4 & 112 & 28.02 & 32.12 & 9.53 & 45.0 & 22.2 & 0.92 \\
\method, ResNet-18 & 8 & 112 & 1.07 & 34.20 & \textbf{9.23} & 41.0 & 24.4 & 0.83 \\
\method, 56 tokens & 8 & 56 & 28.02 & 35.34 & 10.70 & 47.1 & 21.2 & 0.93 \\
\method, 224 tokens & 8 & 224 & 28.02 & 34.08 & 10.48 & 46.1 & 21.7 & 0.93 \\
\textbf{\method, 112 tokens} & 8 & 112 & 28.02 & \textbf{35.67} & 10.61 & 45.9 & 21.8 & 0.93 \\
\bottomrule
\end{tabular}
\end{table*}

\begin{figure*}[t]
    \centering
    \includegraphics[width=\textwidth]{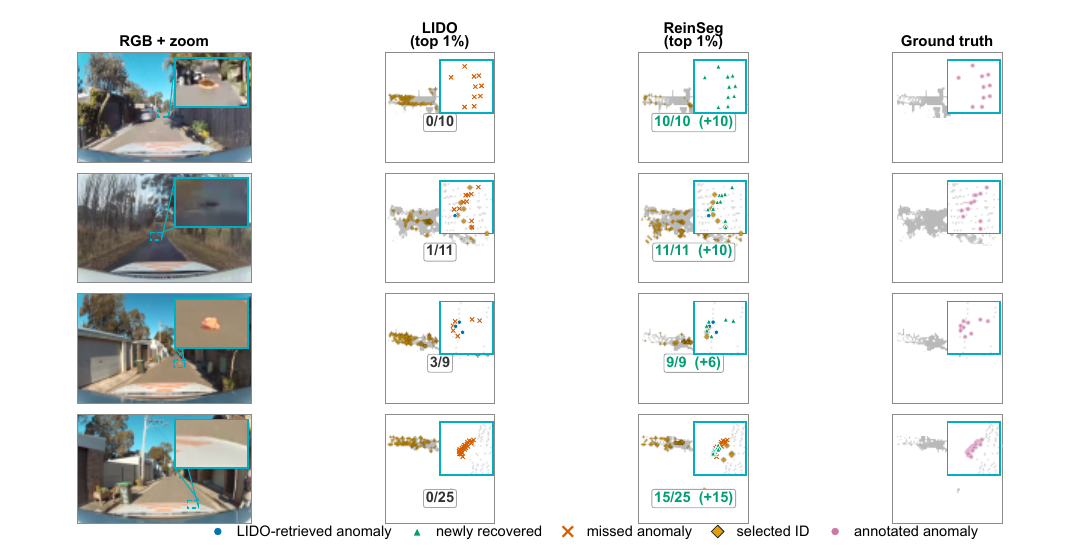}
    \caption{Qualitative comparison at a fixed top-1\% review budget. Each row
    shows a representative case with the RGB frame and three views of the
    evaluated LiDAR scan: LIDO's top-1\% selection, \method's top-1\% selection,
    and ground truth. Cyan locators connect every overview to the same magnified
    hazard region. Blue circles and green triangles denote LIDO-retrieved and
    newly recovered anomaly points, orange crosses mark missed anomalies, amber
    diamonds mark selected ID points, and magenta circles mark annotated
    anomalies. Badges report retrieved/annotated points and the green increment.}
    \label{fig:retrieval-evidence}
\end{figure*}

Figure~\ref{fig:retrieval-evidence} presents representative cases at a fixed
top-1\% review budget. For small, sparsely sampled hazards that are clearly
visible in the camera images, LIDO places only a few of their returns within the
budget. Under the same budget, \method promotes additional returns on the same
hazards while the surrounding road surface stays below the budget, consistent
with the reduced road-surface screening burden.

\section{Conclusion}
\label{sec:conclusion}

We presented \method, a retrieval-guided bounded multimodal refinement
framework for point-wise road anomaly segmentation. Occupied LiDAR BEV cells
retrieve scene-specific context from a surround-view visual memory without
calibration input, and a relative-norm-bounded residual returns this context to
the point features consumed by the frozen LIDO scorer. On STU, \method raises
average precision from 27.54\% to 35.67\%, lowers FPR95 from 34.85\% to 10.61\%,
and reduces the high-recall screening burden by more than a factor of three.
Relative to REL, it also improves AP by 24.99 points and lowers FPR95 by 0.99 points.
Matched interfaces, component isolation, and sensor interventions trace these
gains to retrieved scene evidence and the explicit bound, and show that degraded
cameras never push the system below LiDAR-only performance. Because the retrieval
module consumes no calibration variables and runs at 21.8 frames per second,
the camera path acts as a safe, deployable augmentation of a LiDAR safety
monitor. Bounded retrieval of
visual evidence thus offers a reliable and efficient path to risk-aware 3-D
road-hazard perception in intelligent transportation systems.

\FloatBarrier

\begingroup
\interlinepenalty=10000
\bibliographystyle{IEEEtran}
\bibliography{references}
\endgroup

\end{document}